\documentclass{bmvc2k}

\title{One-shot Face Reenactment}

\addauthor{Yunxuan Zhang}{zhangyunxuan@sensetime.com}{1}
\addauthor{Siwei Zhang}{zhangsiwei@sensetime.com}{1}
\addauthor{Yue He}{heyue@sensetime.com}{1}
\addauthor{Cheng Li}{chengli@sensetime.com}{1}
\addauthor{Chen Change Loy}{ccloy@ntu.edu.sg}{2}
\addauthor{Ziwei Liu}{zwliu@ie.cuhk.edu.hk}{3}
\addinstitution{SenseTime Research}
\addinstitution{Nanyang Technological University}
\addinstitution{The Chinese University of Hong Kong}
\runninghead{ZHANG ET AL.}{One-shot Face Reenactment}

\def\eg{\emph{e.g}\bmvaOneDot}

\usepackage{threeparttable}
\begin{document}

\maketitle

\begin{abstract}
   To enable realistic shape (\eg pose and expression) transfer, existing face reenactment methods rely on a set of target faces for learning subject-specific traits. However, in real-world scenario end-users often only have one target face at hand, rendering existing methods inapplicable. In this work, we bridge this gap by proposing a novel one-shot face reenactment learning framework. Our key insight is that the one-shot learner should be able to disentangle and compose appearance and shape information for effective modeling. Specifically, the target face appearance and the source face shape are first projected into latent spaces with their corresponding encoders. Then these two latent spaces are associated by learning a shared decoder that aggregates multi-level features to produce the final reenactment results. To further improve the synthesizing quality on mustache and hair regions, we additionally propose FusionNet which combines the strengths of our learned decoder and the traditional warping method. Extensive experiments show that our one-shot face reenactment system achieves superior transfer fidelity as well as identity preserving capability than alternatives. More remarkably, our approach trained with only one target image per subject achieves competitive results to those using a set of target images, demonstrating the practical merit of this work. Code\footnote{\url{https://github.com/bj80heyue/Learning\_One\_Shot\_Face\_Reenactment}}, models and an additional set of reenacted faces have been publicly released at the project page\footnote{\url{https://wywu.github.io/projects/ReenactGAN/OneShotReenact.html}}.
\end{abstract}
\vspace{-0.5cm}
\section{Introduction}

\begin{figure}[ht]
\begin{center}
\includegraphics[width=0.95\linewidth]{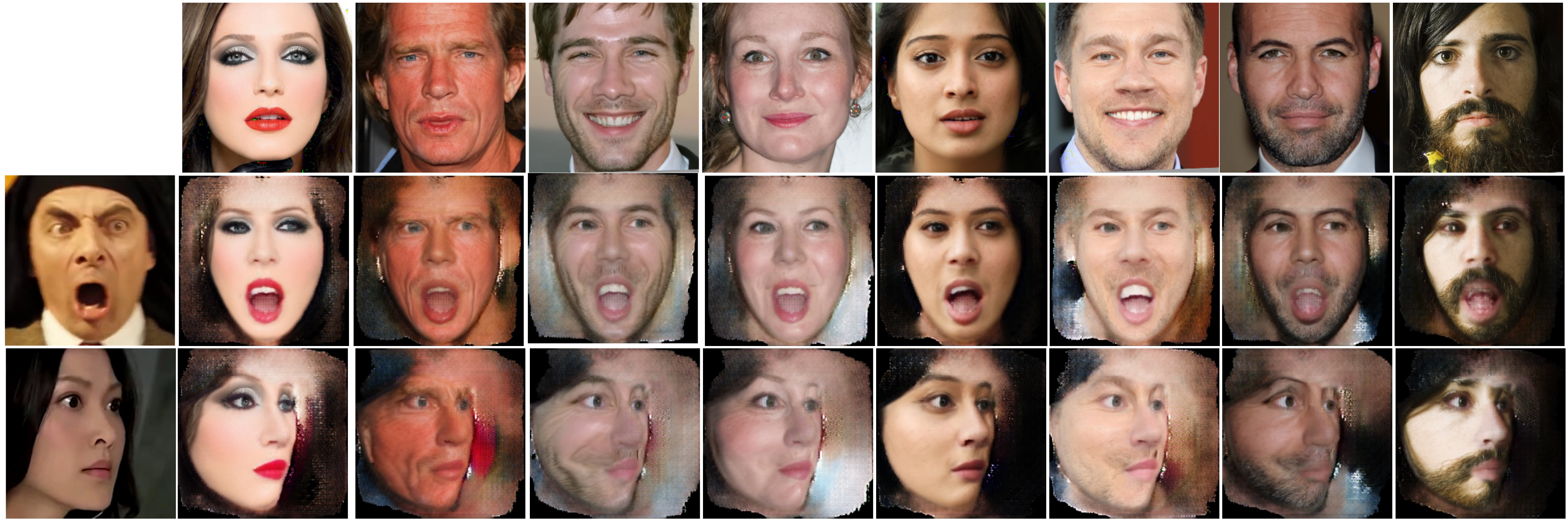}
\vspace{-0.3cm}
\caption{\textbf{The problem of one-shot face reenactment:} With only one reference image input (the top row), our model can reenact under an arbitrary pose guide (the left column). Notice that all result images are generated by one model trained by one-shot data.}
\label{fig:figure1}
\end{center}
\vspace{-0.5cm}
\end{figure}

Face reenactment is an emerging conditional face synthesis task that aims at fulfiling two goals simultaneously: 1) transfer a source face shape (\eg facial expression and pose) to a target face; while 2) preserve the appearance and the identity of the target face.
In recent years, it has attracted enormous research efforts due to its practical values in virtual reality and entertainment industries.

An ideal face reenactment system should be capable of generating a photo-realistic face sequence following the pose and expression from the source sequence when only \textbf{one shot or few shots} of the target face are available.
However, to enable realistic shape (\eg pose and expression) transfer, existing face reenactment methods rely on a set of target faces for learning subject-specific traits.
Some recent works \cite{wu2018reenactgan, kim2018deep, chan2018everybody} relaxed the problem by assuming that the reenactment system can be trained on a relatively long video from the target face. 
For each target face, they train a subject-specific generator using an auto-encoder framework. 
All these methods require a long video from the target and long training time to obtain the target-specific model, limiting their potential applications.

In this work, we propose a novel one-shot face reenactment learning framework to bridge this gap. In our problem setting, we assume that only a single shot is available for each person no matter during training or testing (Figure \ref{fig:figure1}).
Several challenges exist for one-shot face reenactment:
%
1) The appearance of the target person is partial for all views since we only have one reference image from the target person. Synthesizing an image with an arbitrary view with such a limited input constraint is still an open question. 
2) A single image can only cover one kind of expression. The appearance of other expressions (closed eyes, opened mouth) is not available, making the training more challenging. 
3) Identity preserving when only a single target shot is available becomes a hard problem. 

\begin{figure*}[t]
\begin{center}
\includegraphics[width=\linewidth]{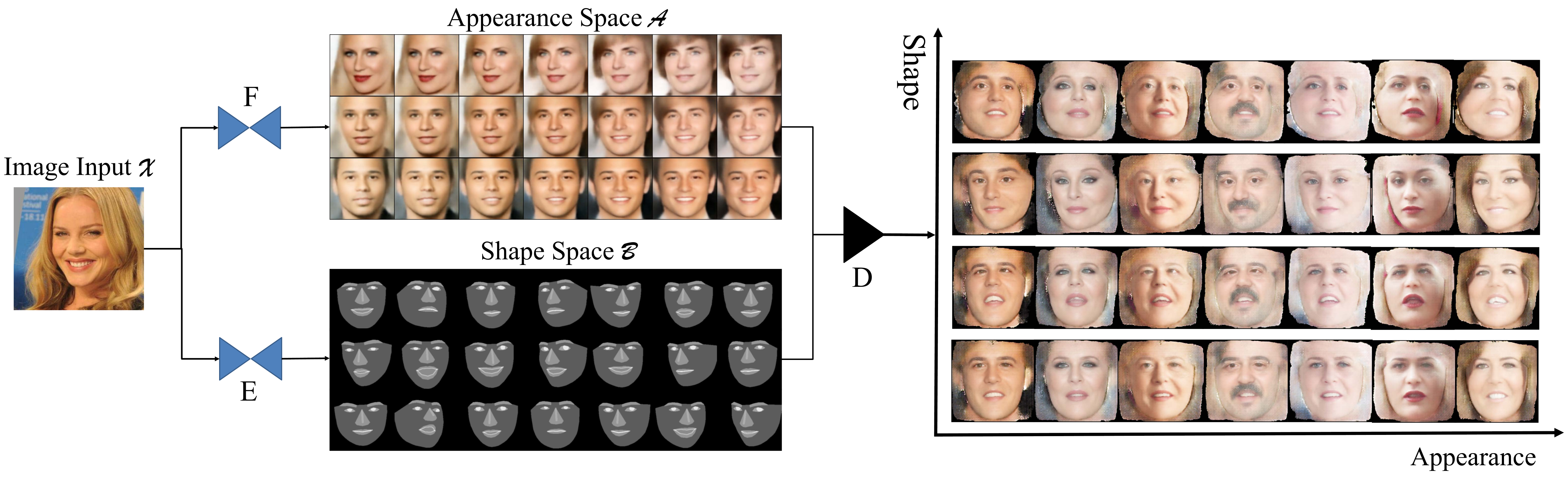}
\vspace{-0.8cm}
\caption{\textbf{Disentangled feature space:} Two separate encoders are adopted to disentangle the appearance and shape spaces, and a joint reenactment decoder composes encoded features into reenactment results. Here the appearance space visualization is reconstructed reference images, and the shape encoder outputs segmentation for pose guides. }
\label{fig:abSpace}
\end{center}
\vspace{-0.8cm}
\end{figure*}

We believe the key to overcoming the aforementioned challenges is the capability of disentangling and composing appearance and shape information.
To this end, the proposed one-shot learner first disentangles the appearance information $a \in \mathcal{A}$ provided by the target face and shape $b \in \mathcal{B}$ provided by source. Then it combines this information by a decoder $D: \mathcal{B} \times \mathcal{A} \rightarrow \mathcal{X}$, which takes full advantage of $a$ so that the model can be trained by one-shot data.
The disentangled feature space is visualized in Figure~\ref{fig:abSpace}.

The identity representation $a$ is critical here. We expect the representation keeps as many facial texture details as possible, but not pose and expression dependent. Achieving this goal is nontrivial: 1) Face representation learned by some difficult tasks like face verification seems to be a good candidate. However, our experiments show that it cannot retain low-level texture information well. 2) We require a representation that can support face reconstruction but exclude the facial pose and expression of the input. 3) There exists a natural gap between training and testing. During the testing phase of the reenactment, the appearance $a$ and shape $b$ belong to different identities. However, it is infeasible to obtain face pairs with the same head pose and expression across different identities for training.

We found that inappropriate encoding of appearance can hurt model performance considerably, \eg, losing facial details such as eye colors or lip thickness due to insufficient information carried by the encoder's output, or identity shift due to facial structure changes. To supervise our shape-independent face appearance representation, we link two branches over the face appearance decoder: one for appearance reconstruction and the other one for the reenactment. The two branches are trained jointly. These two branches both take the output of the appearance encoder as input. In other words, an auto-encoder is trained for appearance encoding and reconstruction, and a reenactment decoder is trained in parallel with the appearance decoder to transfer pose to the target face. 

To alleviate appearance and identity information loss during the reenactment process, we propose a novel strategy that inserts feature maps of different scales from the appearance decoder into multiple intermediate layers of the reenactment decoder to make full use of the appearance information. Besides, we preserve shape semantic information by inserting SPADE module \cite{park2019semantic}\footnote{SPADE is equivalent to Spatial Feature Transform (SFT)\cite{wang2018recovering} with normalization layers.} into the reenactment decoder in a multi-scale manner to achieve a semantically adaptive decoder. Additionally, we propose a FusionNet to improve the performance on mustache and facial texture generation.
Results of our one-shot reenactment model are shown in Figure \ref{fig:figure1}.

The contributions of this work can be summarized as follows:
1) We study a new problem ``one-shot face reenactment''. This study offers a practical framework for real reenactment scenarios with the notion of one-shot training and test settings.
2) We propose a novel one-shot face reenactment learning framework. It disentangles and composes appearance and shape information for effective modeling. We additionally design FusionNet that combines the strengths of synthesis-based method and warping-based approach to better preserve texture information.
Extensive experiments validate our framework that it can be trained with one-shot data while still able to reenact realistic face sequences with just only one image from the target person. More remarkably, the results of the proposed approach outperform state-of-the-art single-image face generating methods and are even competitive with subject-specific methods.

\vspace{-0.5cm}
\section{Related Work}
\vspace{-0.2cm}
\noindent \textbf{Face Reenactment.} Here we roughly divide the existing methods for face reenactment into two categories: model-based methods \cite{thies2016face2face, kim2018deep, cheng20093d, thies2015real, vlasic2005face, Dale2011Video, garrido2015vdub} and feed-forward-based methods \cite{kim2018deep, wu2018reenactgan, wiles2018x2face, jin2017cyclegan, xu2017Face, zhou2019talking}. 
The model-based methods typically have three steps: 1) Capture face movement by a face tracking or using optical flow on either RGB or RGBD camera. 2) Fit the movement of the face into a parametric space or a 3D model. 3) Once the model is fitted, the final stage is to re-render a new video. 
The feed-forward method has a similar pipeline but replaces the third re-render step with a neural network generator. 
For example, Deep video portrait \cite{kim2018deep} takes 3D rendering results, illumination and the expression as inputs to generate a near-real quality image. ReenactGAN \cite{wu2018reenactgan} takes only the boundary information as input to achieve a competitive result. 
However, these works demand a large amount of training data for subject-specific generation. Besides, a separate training process is required for each single target face.  

\noindent \textbf{One-shot Generative Model.} 
There exist facial attributes manipulation methods \cite{choi2018stargan,chen2018facelet, yin2019geogan, chang2018pairedcyclegan, yang2018learning,pumarola2018ganimation} that support single image input. 
However, they only alter the appearance of facial components and face difficulty in handling large shape and pose changes. 
Meanwhile, there are one-shot models that generate images under the guidance of texture and semantic information, such as full-body generator~\cite{ma2017pose, ma2018disentangled, balakrishnan2018synthesizing, li2019dense, esser2018variational,ma2017pose,zhang2018joint, tran2017disentangled} and the street view generator~\cite{park2019semantic}. These methods are confronted with identity shift or coarse details of face when adopted to generate face image directly.
In addition, prior studies fail to enable one-shot face reenactment based on an explicit continuous pose representation, due to the challenges in large shape changes, the complexity of pose and expression, and the need for realistic facial details.

\vspace{-0.5cm}
\section{Approach}
\label{sec:approach}
\vspace{-0.2cm}

\begin{figure*}[ht]
\begin{center}
\includegraphics[width=0.80\linewidth]{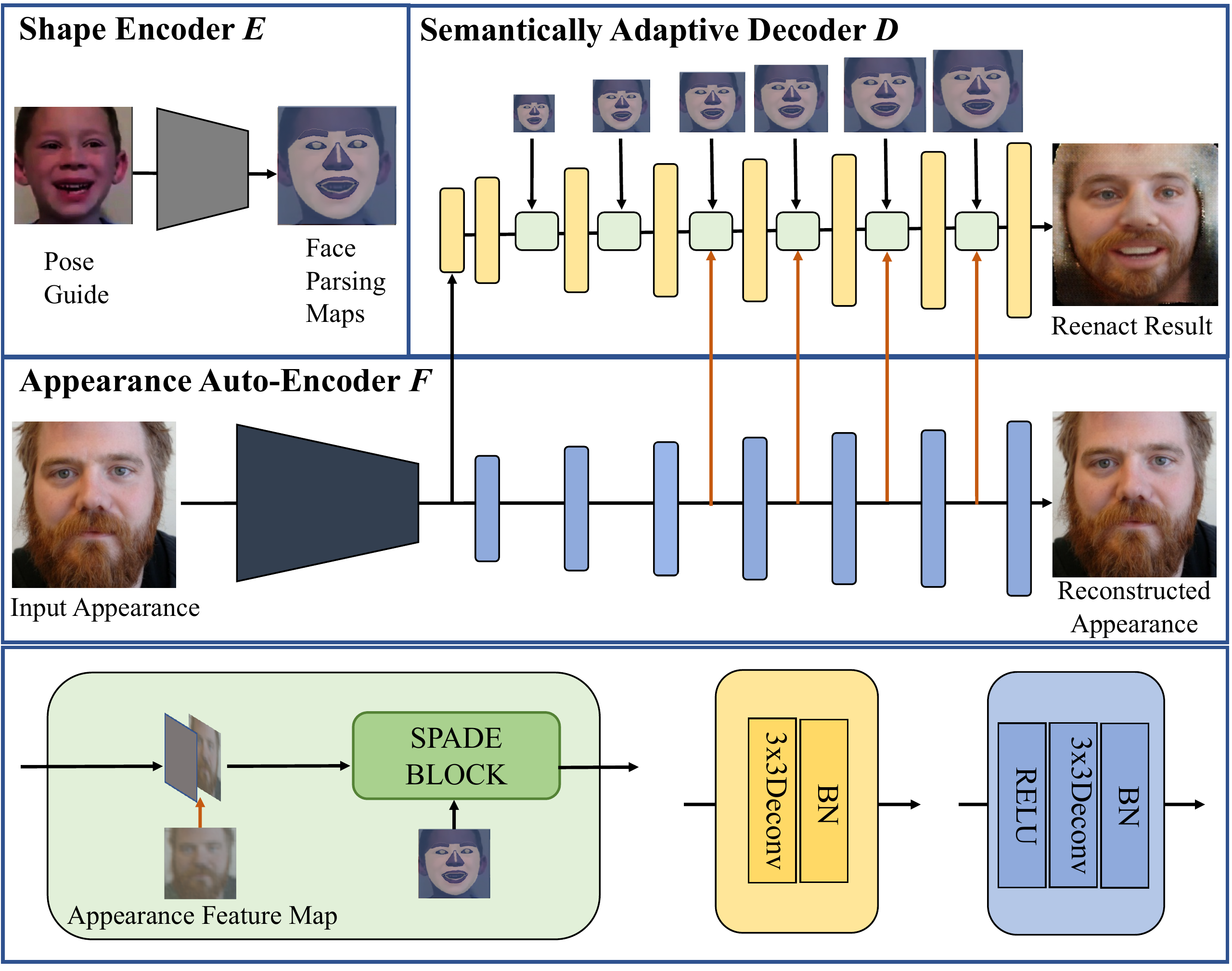}
\vspace{-0.3cm}
\caption{\textbf{An overview of the proposed one-shot reenactment model:} 
Given a target image (the input appearance) and a source image (the pose guide) as input, we extract appearance features of the target image with $F$ and feed the source image to $E$ to generate face parsing maps simultaneously.
The decoder $D$ composes shape and appearance information to generate the reenactment result, which preserves the reference's texture and the pose guide's shape.
In order to improve the reenactment result, we insert SPADE residual blocks \cite{park2019semantic} to preserve spatial information and concatenate the feature maps from the face reconstruct decoder to $D$ in a multi-scale manner.
The $D$ and $F$ are trained simultaneously while the $E$ is pretrained.
}
\label{fig:one_shot_overview}
\end{center}
\vspace{-0.6cm}
\end{figure*}

\noindent\textbf{Overview.}
Given a source face, $x_s$, which provides the pose guidance, and a target face, $x_t$, which provides the reference appearance, the goal is to transfer the expression and pose of $x_s$ to $x_t$.
Specifically, for any input face $x \in \mathcal{X}$, we wish to first encode it into the Cartesian product of appearance representation $a \in \mathcal{A}$ and the shape representation $b \in \mathcal{B}$ through an encoder $F: \mathcal{X} \rightarrow \mathcal{A}$ and $E: \mathcal{X} \rightarrow \mathcal{B}$.
Then we need a general decoder $D: \mathcal{B} \times \mathcal{A} \rightarrow \mathcal{X}$ to map the appearance and shape from the latent embedding space back to the face space. The reenacted face is generated by computing $D(F(x_t), E(x_s))$.
Figure \ref{fig:abSpace} illustrates the three spaces $\mathcal{A}$, $\mathcal{B}$ and $\mathcal{X}$ learned by our framework.


\vspace{-0.3cm}
\subsection{Disentangle-and-Compose Framework}

Figure \ref{fig:one_shot_overview} shows the overview of the proposed one-shot reenactment model. 
We first disentangle the shape and appearance by a face parsing model ($E$) and an appearance auto-encoder ($F$), respectively.
Then we progressively compose the source expression and the target identity into a realistic face. In particular, the decoder $D$ is composed of multiple intermediate blocks (a yellow box and a green box in Figure \ref{fig:one_shot_overview} form a block). Each of the blocks concatenates feature maps from $F$ that captures the appearance information with that of the corresponding layer in $D$. Each block further uses the SPADE block \cite{park2019semantic}, which takes face parsing maps as input, to transfer the spatial information originated from the source face. To make full use of the encoding information, we insert intermediate blocks of different scales, taking multi-scale appearance encodings and face parsing maps as inputs.

\noindent\textbf{Shape Encoder ($E$).} 
Here we apply the same boundary encoder as \cite{wu2018reenactgan} and \cite{wu2018look}. 
The boundary encoder, which is trained on WFLW dataset, maps a cropped face into a 15-channel heatmap, corresponding to different parts of the face. 
Then we fill different colors in different facial parts (nose, eyes, etc.) of the boundary heatmap to produce the face parsing maps.
Furthermore, we add two additional gaze channels to model the gaze movement of the face. The gaze channels are trained on EOTT dataset \cite{papoutsaki2018eye}.
The shape encoder is pretrained, and frozen during the training of the following two modules, namely Appearance Auto-Encoder ($F$), and Semantically Adaptive Decoder ($D$).

\noindent\textbf{Appearance Auto-Encoder ($F$).}
Recall that appearance encoding is crucial for one-shot face reenactment.
The appearance encoder needs to capture sufficient information, including identity information and local facial details, from just a single target image.
Rather than concatenating the reference face to intermediate layers of $D$ directly, we adopt the feature maps from an appearance auto-encoder, $F$, which takes the reference face $x_t$ as input.
In comparison to a fixed appearance encoder, we found that the learned model can better preserve low-level textures and identity information.

The appearance auto-encoder includes a decoder (the blue blocks in Figure \ref{fig:one_shot_overview}) that aims to reconstruct the target image from the latent feature maps. To better preserve the identity during face reenactment, we extract multi-layer features from the decoder and concatenate them with the corresponding features in the Semantically Adaptive Decoder, $D$.
Note that the appearance encoder should stay independent from the shape.
To achieve this goal, instead of using the same image for both the appearance reference and pose guide, we replace the face contour of the pose guide in the face parsing map with the face contour of another randomly sampled person during the training stage, while keeping inner facial parts unchanged, to encourage the $F$ to ignore shape information as much as possible.

\noindent\textbf{Semantically Adaptive Decoder ($D$).}
\label{semantically adaptive decoder}
%
A U-Net is frequently used in generation tasks \cite{karras2017progressive,wang2018pix2pixHD,zhu2017unpaired} as its skip-connect structure can alleviate information loss. Normalization layers such as BatchNorm and InstanceNorm are also widely used to escalate model convergence. However, as shown in \cite{park2019semantic}, these commonly used normalization layers will uniform the spatial information in feature maps and cause semantic information loss.
Inspired by \cite{park2019semantic}, instead of stacking conventional blocks that include convolution and normalization layers to encode semantic information, we insert multi-scales SPADE blocks into the decoder $D$, to transmit spatial information from the face parsing map into the decoder directly by estimating normalize parameters from the parsing. As such, the normalization parameters spatially correlate to the face parsing, therefore eliminating the need for an encoder in a U-Net structure and leading to a network with a smaller size.
The semantically adaptive decoder $D$ is trained together with the appearance auto-encoder $F$ in an end-to-end manner.

\vspace{-0.3cm}
\subsection{FusionNet}\label{sec:FusionNet}
\vspace{-0.2cm}
To eliminate disturbance from the background and to speed up convergence, the proposed face reenactment model focuses on inner facial components.
Observing the results of our model in Figure \ref{fig:abSpace}, the inner facial components are inherently governed by the given face parsing. 
While this method is effective, the model still faces problems in generating facial details such as mustache and wrinkle.
We note that warping-based methods \cite{liu2017voxelflow, yeh2016semantic} has an advantage in generating facial details by warping pixels from the original image. 
Consequently, we propose the FusionNet, which takes the result of our model and the traditional warping result \cite{facewarp} as inputs, and generate a mask to fuse two generated images. 

\begin{figure}[ht]
\vspace{-5pt}
\begin{center}
\includegraphics[width=0.8\linewidth]{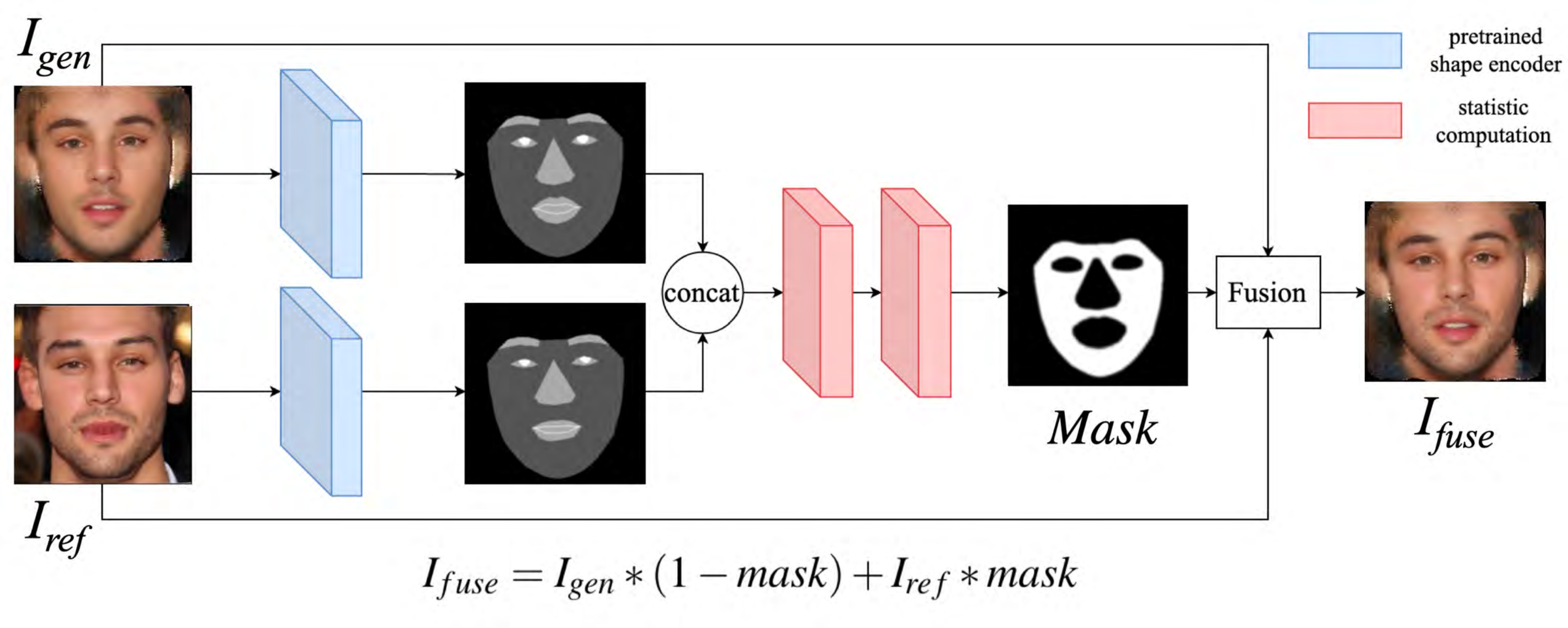}
\vspace{-8pt}
\caption{\textbf{FusionNet:} We propose FusionNet to leverage the face generated from multiple sources. Here we fuse the results from classical warping, which generates better textures, and works well on small pose changes; and our model, which is able to handle large pose changes and exaggerated facial actions. }
\label{fig:fusionNet}
\end{center}
\vspace{-1.0cm}
\end{figure}

\subsection{Learning}
\vspace{-0.2cm}
We define a combined loss for all the components and optimize them jointly, where the $\lambda$ and $\alpha$ are the balancing weights,
$L_{app\_recons}$ is the L1 loss between input appearance and the reconstructed appearance, $L_{reconstruct}$ is the L1 loss between the reenactment result and the input appearance after face contour resampling, $L_{gan}$ is the multi-scale GAN loss~\cite{wang2018pix2pixHD}, $L_{perceptual}$ is the perceptual loss~\cite{wang2018pix2pixHD} between the reenactment result and input appearance image after the face contour resampling,
and $L_{id}$ is the L1 loss between verification features of the reenactment result and the input appearance produced by a verification model trained on CelebA dataset \cite{liu2015deep}, where $\lambda = 25$, $\alpha_r = 25$ and $\alpha_p, \alpha_g, \alpha_i = 1$ are used in the experiments:
\begin{equation} \label{eq:combine_loss}
\begin{aligned}
    L_{total} = &L_{reenact} + \lambda L_{app\_recons}, \\
    L_{reenact} = &\alpha_{r}L_{reconstruct} + \alpha_{p}L_{perceptual} + \alpha_{g}L_{GAN} + \alpha_{i}L_{id},
\end{aligned}
\end{equation}

\vspace{-0.5cm}
\section{Experiments}
\vspace{-0.2cm}
We compare our one-shot face reenactment framework with two types of methods: 1) one-shot image generating methods (Sec.~\ref{sec:singleImageMethod}). 2) target-specific face reenactment methods (Sec. \ref{sec:targetSpecificMethod}). 
In Sec. \ref{sec:quantitativeComparison}, we quantitatively evaluate the quality of the generated image by computing the facial action and pose consistency between generated results and the condition pose. In addition, we also measure different models' capability in preserving the target's identity. 

\noindent\textbf{Implementation Details.}
We implement our model using PyTorch.
The learning rates for the generator and discriminator are
set to 1e-4 and 5e-5. 
We use ADAM \cite{kingma2014adam} as the optimizer and set $\beta_1 = 0, \beta_2 = 0.999$.

\noindent\textbf{Network Architecture.} 
There are eight downsampling blocks in the appearance encoder, with each block containing [convolution, instanceNorm, LeakyReLU] and these convolutional layers have 64, 128, 256, 512, 1024, 1024, 1024, 1024 channels with filter size of 4x4. For the appearance decoder, there are seven upsampling blocks with each block containing [deconvolution, instanceNorm, ReLU] and the deconvolutional layers have 1024, 1024, 1024, 1024, 512, 256, 128 channels with filter size of 4x4. The semantically adaptive decoder includes eight upsample modules, with the last four modules containing [ResBlkSPADE, convolution, upsample], and the first four modules only containing convolution and upsample layers. Here ResBlkSPADE refers to the SPADE residual blocks in \cite{park2019semantic}.

\noindent\textbf{Training Set.} 
For the training set, we use a subset of the CelebA-HQ \cite{karras2017progressive} with 20k images. The high-quality of face images in CelebA-HQ helps the model to learn more details. 
The faces are first detected by Faster R-CNN \cite{ren2015faster}, then rigidly transformed into a mean shape and cropped to $256 \times 256$.

\noindent
\textbf{Testing Set.} 
For pose guide data during testing, we sample 100 expressions that are unseen in the training set for each target face and cover different expressions and head pose including yaw, pitch and roll.
Considering the gaps between different datasets and in order to assess the general capacity of our one-shot reenact model, we prepare three kinds of test datasets, each of which contains 500 images of different identities:
1) Target data from the same source, selected from CelebA-HQ with unseen identities during training.
2) Target data from a different source, selected from another high-quality face image dataset FFHQ \cite{karras2019style}, with similar distributions of gender and ethnicity.
3) Target data in the wild, selected from a real-world dataset RAF-DB \cite{li2017reliable}.

\vspace{-0.4cm}
\subsection{Comparison with Single-Image Methods}
\label{sec:singleImageMethod}

\begin{figure*}[!ht]
\begin{center}
\includegraphics[width=1\linewidth]{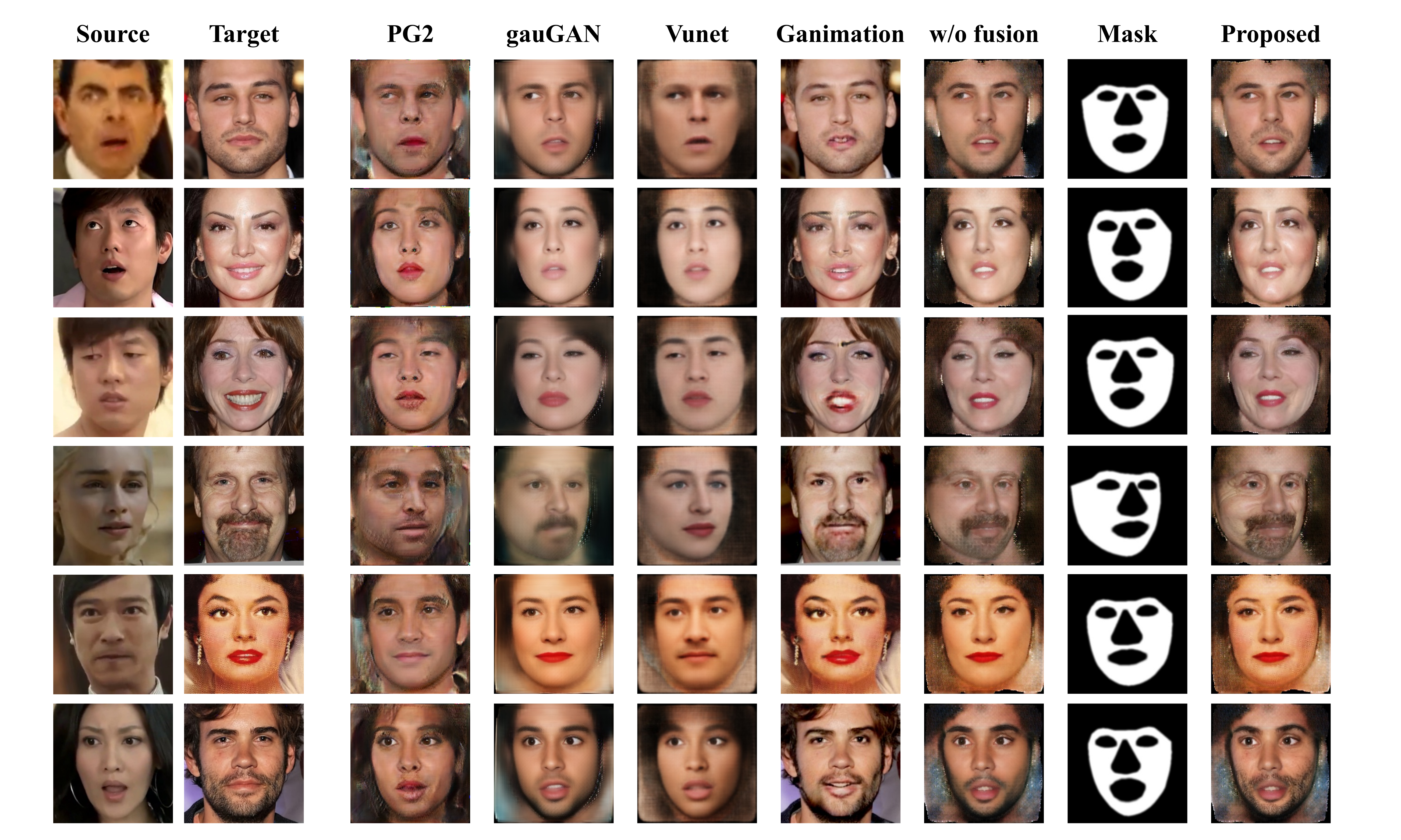}
\vspace{-0.8cm}
\caption{\textbf{Qualitative comparison with state-of-the-art single-image methods:} Our proposed framework generates faces effectively under larger pose changes compared to GANimation, with better identity consistency compared to PG2 \cite{ma2017pose}, gauGAN \cite{park2019semantic} and VU-Net \cite{esser2018variational}. }
\label{fig:compareSingleImageMethod}
\end{center}
\vspace{-0.8cm}
\end{figure*}

We first compare our framework with several state-of-the-art methods that support single-input generation. 
Since there are few single-input face reenactment models, we also compare our model with single-input generation models in other domains such as human pose, cityscapes and landscapes and train these model with our training data: 
1) \textbf{GANimation} \cite{pumarola2018ganimation}, a conditional GAN based method which generates faces conditioned on facial action unit. 
2) \textbf{PG2} \cite{ma2017pose}, a pose guided person generation network
that can synthesize person images in arbitrary poses, based on a reference of that person and a novel pose guide. 
3) \textbf{gauGAN} \cite{park2019semantic}, which allows user control over both semantic and style for synthesizing images with the target semantic map and style.
4) \textbf{VU-Net} \cite{esser2018variational}, which synthesizes images of objects conditioned on shape information by using VAE.
5) The proposed reenact model without FusionNet.

Figure \ref{fig:compareSingleImageMethod} shows the result of all baselines and the proposed framework. 
GANimation performs well on inner face components but with mostly static pose, and it cannot handle all expressions. 
Despite the high-quality image generation in other domains, PG2 and gauGAN are confronted with the identity shift problem.
Vanilla VUNet generates blurred inner face, and faces the identity shift problem as well.
Our proposed method can simultaneously modify the expression of the inner face and the whole head pose while still preserving the face identity.

\vspace{-0.2cm}
\subsection{Comparison with Target-Specific Methods} \label{sec:targetSpecificMethod}
\vspace{-0.1cm}

\begin{figure*}[ht]
\begin{center}
\includegraphics[width=\linewidth]{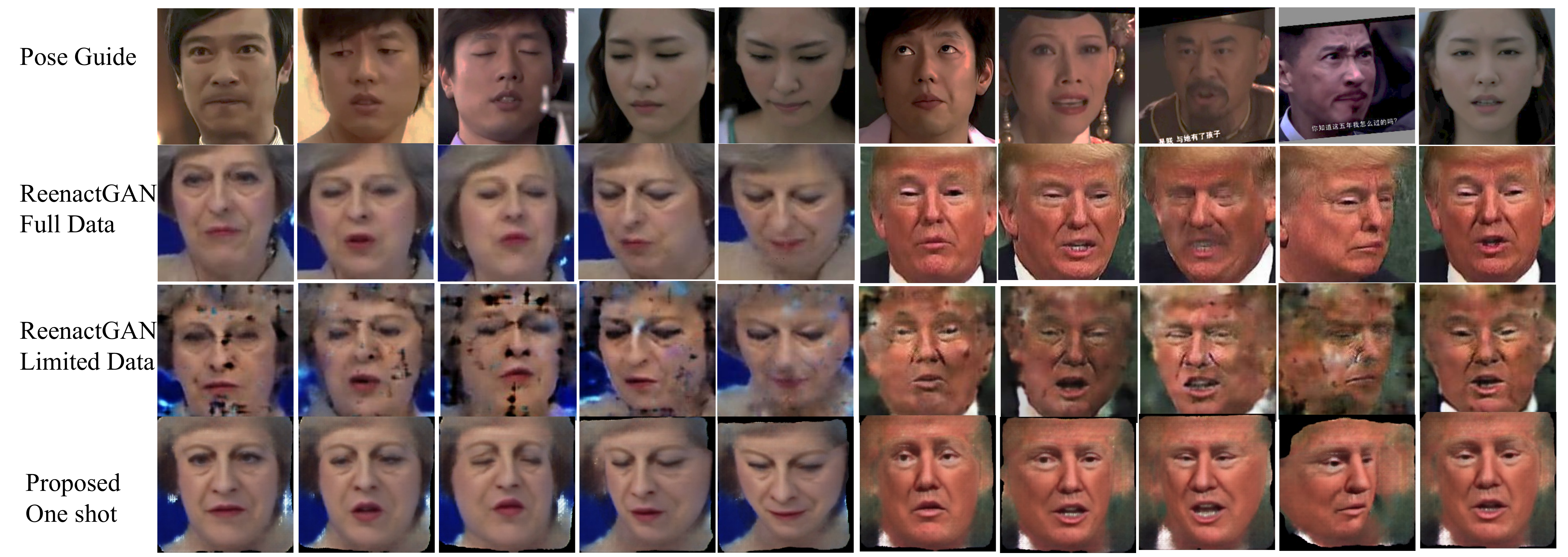}
\caption{\textbf{Qualitative comparison with target-specific methods:} ReenactGAN requires adequate data to train target-specific model for each target and fails when the training data decreases significantly.
Contrast to that, with a few or even only one reference image, our framework is able to generate competitive results. }
\label{fig:compareTargetSpecific}
\end{center}
\vspace{-0.4cm}
\end{figure*}

Here we compare our model with the state-of-the-art algorithm ReenactGAN~\cite{wu2018reenactgan}.
In target-specific setting, the model can be trained on a video from the target person $\{x^T_1, \ldots , x^T_M \}$, which usually provides enough appearance information of the target person under different expressions.
But in practice, it usually requires expensive computation resources to train the target-specific model as well as extra efforts to collect large amounts of high-quality data to reenact a new person.
In contrast, our one-shot framework enables users to generate reenactment video with a single target photo input without extra training.
Figure \ref{fig:compareTargetSpecific} shows that the target-specific model fails when given a limited number of target data while our proposed method achieves competitive results taking only one image of the target person.

\vspace{-0.3cm}
\subsection{Quantitative Results} \label{sec:quantitativeComparison}

In this section, we quantitatively compare our method and the aforementioned methods from three aspects: 1) Expression consistency of inner facial components. 2) The consistency of head pose between the source video and the generated result. 3) The capability of preserving identity. 

\noindent
\textbf{Facial AU Consistency.} Facial action unit (AU) consistency is widely applied in studies on face generation \cite{pumarola2018ganimation, wu2018reenactgan}. Here we evaluate the consistency between the generated sequence and the source sequence following the protocol from \cite{wu2018reenactgan}.

\noindent
\textbf{Head Pose Consistency.} Besides the consistency of AU, we also evaluate the consistency of head motion. We apply a head pose estimator model and measure the consistency of row, pitch, yaw angles between generated results and the source sequence. We report the average of mean absolute error of yaw, pitch, roll in degree.

\noindent
\textbf{Identity Preserving.} Instead of using the perception loss, we evaluate identity preserving capability by measuring the recognition accuracy with a fixed model \cite{wang2017residual}. 
We compute the overall accuracy by judging if each generated face has the same identity with the reference face.

\begin{table}[t]
\footnotesize
\vskip 0.1cm
\centering
\caption{\textbf{Quantitative comparison:} We evaluate each method's consistency of facial action unit (AU), head pose, and the ability in preserving identity. Here we divide the testing data into three groups: 1) identical source. 2) cross source (different source but with similar distributions of gender and ethnicity). 3) in the wild. For the reported AU consistency and ID preserving metrics, higher is better, whereas for the pose consistency metric lower is better. 
 }
\vskip 0.1cm
\begin{threeparttable}
\begin{tabular}{c|ccc|ccc|ccc}
   \hline
   & \multicolumn{3}{|c|}{same source} & \multicolumn{3}{|c|}{cross source} & \multicolumn{3}{|c}{in the wild} \\\hline
  Method &  AU \% & Pose & Id\% & AU \% & Pose & Id\% & AU \% & Pose & Id\% \\
  \hline\hline
  Reference    & 0.00 & 7.49 & - & 0 & 7.77 &  - & 0 & 7.51 & -\\
  GANimation \cite{pumarola2018ganimation} \tnote{*}  & 18.2 & 7.49 & \textcolor{blue}{99.6} & 17.1 & 7.77 & \textcolor{blue}{99.5} & 13.5 & 7.50 & \textcolor{blue}{99.3}   \\
  VUNet \cite{esser2018variational} & 80.2 & 3.04 & 46.1 & 79.4 & 3.03 & 47.7  & 79.6 & 3.03 & 52.2    \\
  PG2 \cite{ma2017pose} & \textcolor{blue}{82.7}  & 3.19 & 52.3 & \textcolor{blue}{82.6}  & 3.18 & 53.7  & \textcolor{blue}{82.3} & 3.18 & 51.9  \\
  GauGAN \cite{park2019semantic} & \textcolor{red}{81.1} & 2.98 & 79.3   & 79.72 & 2.98 & 75.3 & 79.7 & 3.00 & 78.7   \\ \hline
  Ours w/o Fusion     & 80.0 & \textcolor{red}{2.79} & 89.1  & \textcolor{red}{80.22} & \textcolor{red}{2.84} & 88.2 & \textcolor{red}{80.0} & \textcolor{red}{2.91} & 90.2    \\
  Ours     & 75.1 & \textcolor{blue}{2.72} & \textcolor{red}{98.2}  & 70.9 & \textcolor{blue}{2.74} &  \textcolor{red}{98.5}  & 71.1 & \textcolor{blue}{2.63} & \textcolor{red}{98.3} \\
  \hline
 \end{tabular}
 \begin{tablenotes}
 \item[*] Different from our training setting, GANimation is supervised by action-unit labels.  
 \end{tablenotes}
     \end{threeparttable}
\label{tab:quantitativeResult}
\end{table}     
 
We evaluate the aforementioned methods with three metrics. We also prepare a baseline that treats reference faces as output, which should have the lowest consistency on AU or pose consistency and no identity shifting. 
Table \ref{tab:quantitativeResult} summarizes all quantitative results; we can observe that: 1) The identity preserving ability of our proposed framework overrides the state-of-the-art one-shot methods of other image domain and is competitive with GANimation. 
    2) Our proposed framework performs well in AU and pose consistency while GANimation has frozen pose and cannot handle all expressions. 
    3) Our proposed model has an overall high performance on different data.
    4) The FusionNet improves identity preserving significantly but with a little drop in AU consistency, because the FusionNet focuses on the quality of face rather than the expression. Note that the PG2 is trained by pair data which has better supervision in expression, so it has the best AU consistency. \\

\vspace{-0.2cm}
\noindent
\textbf{Ablation Study.}
 We investigate the contribution of the concatenation between the appearance decoder and spatially adaptive decoder, and the influence of $\lambda$, the appearance reconstruction loss weight. 
Table \ref{tab:ablation study} provides the ablation study results in terms of identity preserving metric values. We can see that concatenation boosts identity preservation performance significantly, and an overly large $\lambda$ will cause overfitting on the appearance reconstruction on training data.
\begin{table}[h]
\footnotesize
\vskip 0.1cm
\centering
\caption{\textbf{Ablation study:} We evaluate the identity preserving performance under different appearance reconstruction settings on the same source test data.}
\vskip 0.1cm
\begin{tabular}{c|c|c|c|c|c}
\hline
& No concat & $\lambda=3$ & $\lambda=10$ & $\lambda=25$ & $\lambda=50$ \\
\hline\hline
Id\% & 77.7 & 84.9 & 86.9  & \textbf{89.1} & 85.2 \\
\hline
\end{tabular}
\label{tab:ablation study}
\end{table}

\noindent
\textbf{$k$-shot Analysis.}
We also inspect how the one-shot model will perform in a few-shot situation. So we collect an additional few-shot dataset that is extracted from several films. There are a total of 32 identities and each identity has seven images with a different expression and pose. Then we conduct 3-shot and 5-shot experiments to examine if more data can improve the performance. Table \ref{tab:few_shot} shows the identity preserving performance under few-shot data and our model performs better when more data are available. 

\begin{table}[h]
\footnotesize
\vskip 0.1cm
\centering
\caption{\textbf{Few shot:} We evaluate the identity preserving performance under few-shot data.}
\vskip 0.1cm
\begin{tabular}{c|c|c|c}
\hline
& one-shot & 3-shot & 5-shot \\
\hline\hline
Id\% & 97.2 & 99.3 & \textbf{99.4}  \\
\hline
\end{tabular}
\label{tab:few_shot}
\end{table}

\section{Conclusion}

We have presented a one-shot face reenactment model that takes only one image input from the target face and is capable of transferring any source face's expression to the target face. 
Specifically, we leveraged disentangle learning to model the appearance and the expression space independently and then compose this information into reenactment results.
We also introduced a FusionNet to further leverage the generated result and the result from traditional warping method and improve the performance on texture and mustache.
Our approach trained with only one target image per subject achieves competitive results to those using a set of target images, demonstrating the practical merit of this work. \\

\noindent\textbf{Acknowledgements.}
We would like to thank Wenyan Wu and Quan Wang for their exceptional support. This work is supported by SenseTime Research, Singapore MOE AcRF Tier 1 (M4012082.020), NTU SUG, and NTU NAP.

\bibliography{egbib}

\end{document}